\documentclass[10pt,twocolumn,letterpaper]{article}

\usepackage{iccv}
\usepackage{times}
\usepackage{epsfig}
\usepackage{graphicx}
\usepackage{amsmath}
\usepackage{amssymb}
\usepackage{arydshln}

\usepackage[pagebackref=true,breaklinks=true,letterpaper=true,colorlinks,bookmarks=false]{hyperref}

 \iccvfinalcopy 

\def\iccvPaperID{****} 

\ificcvfinal\fi

\begin{document}

\title{The challenge of representation learning: Improved accuracy in deep vision models does not come with better predictions
of perceptual similarity}

\author{Fritz G\"unther\\
Humboldt-Universit\"at zu Berlin\\
Unter den Linden 6, 10117 Berlin, Germany\\
{\tt\small fritz.guenther@hu-berlin.de}
\and
Marco Marelli\\
University of Milano-Bicocca\\
Piazza dell'Ateneo Nuovo 1, 20126 Milano, Italy\\
{\tt\small marco.marelli@unimib.it}
\and
Marco Alessandro Petilli\\
University of Milano-Bicocca\\
Piazza dell'Ateneo Nuovo 1, 20126 Milano, Italy\\
{\tt\small marco.petilli@unimib.it}
}

\maketitle
\ificcvfinal\thispagestyle{empty}\fi

\begin{abstract}
   Over the last years, advancements in deep learning models for computer vision have led to a dramatic improvement in their image classification accuracy. However, models with a higher accuracy in the task they were trained on do not necessarily develop better image representations that allow them to also perform better in other tasks they were not trained on. In order to investigate the representation learning capabilities of prominent high-performing computer vision models, we investigated how well they capture various indices of perceptual similarity from large-scale behavioral datasets. We find that higher image classification accuracy rates are not associated with a better performance on these datasets, and in fact we observe no improvement in performance since GoogLeNet (released 2015) and VGG-M (released 2014). We speculate that more accurate classification may result from hyper-engineering towards very fine-grained distinctions between highly similar classes, which does not incentivize the models to capture overall perceptual similarities.    
\end{abstract}

\section{Introduction}

Over the last decade, following the seminal work by Krizhevsky, Sutskever, and Hinton \cite{Krizhevsky2012}, computer vision models based on deep neural network architectures have become increasingly powerful, and nowadays achieve very high levels of performance \cite{Byerly2022,Szegedy2017}. This performance is typically assessed on the very task used in model training, most often as the accuracy in image classification (using measures such as top-1 error or top-5 error \cite{Russakovsky2015}).

As these models achieve higher and higher performance in such scenarios, they also tend to become increasingly sophisticated and complex in terms of model architecture and the numbers of parameters to be estimated. However, this additional complexity does not necessarily imply that these models \emph{generally} perform better, also on domains they are \emph{not} trained on: such an approach runs the risk of having systems that are over-optimized for a particular (set of) tasks, without gaining much in terms of transfer and generalizability \cite{TransferLearning}.

These aspects play an important role in machine learning, often discussed under the label of \emph{representation learning} \cite{TransferLearning}). However, the point is even more relevant when these systems are used as general-level vision models for research purposes. In that respect, an emerging line of research in the domains of  computational neuroscience and cognitive science has started to investigate and employ computer vision models (originally designed and trained for image classification)  as models for human visual representation and processing, with very promising results from recent studies \cite{BattledayReview,vispa}. These works also provide us with rich, large-scale datasets of human behavioral data that allow us to investigate to which extent current computer-vision models can serve as general-level vision models, with much wider scientific applications than being pure image classifiers \cite{Cichy2019,Kriegeskorte2015,Lindsay2021}. Following these developments, in the present study, we will systematically examine which models perform best when tested against a battery of behavioral datasets, and if such models also turn out to be the most complex and best-performing image classifiers.

\section{Related Work}

In the development of language models, human behavioral data have long been established as a gold standard for model evaluation (e.g. \cite{Baronipredict}). The most prominent example are ratings of word similarity, with widely-used datasets such as WordSim353 \cite{Finkelstein2001}, SimLex999 \cite{Simlex}, or MEN \cite{Bruni2014}.

Analogously, ratings of image similarity are widely employed to evaluate and compare the performance of computer vision models. This includes pairs of different naturalistic images  \cite{Hebart2020,Jozwik2017,Peterson2018}, as well as comparisons between real images and their distorted versions \cite{Zhang2018}. In a recent study, Roads and Love collected similarity ratings for a very large collection of 50,000 ImageNet images, which were not only used for evaluation but also to enrich computer vision with participant-sourced information \cite{Roads2021}.

More recently, G\"unther al al.  \cite{vispa} released a collection of several large-scale data sets, comprising rating data as well as on-line processing data in the form of response times, which were used to evaluate a VGG-based vision model \cite{Chatfield2014}. These will constitute the gold standard datasets for our present study, where we systematically evaluate the performance of a wide range of models against data that are cognitively relevant, but relatively atypical for the computer vision domain, and far from the tasks on which systems are typically optimized.

\section{Datasets}

\begin{figure}[ht]

\textbf{[DIS]}

\includegraphics[width=\linewidth]{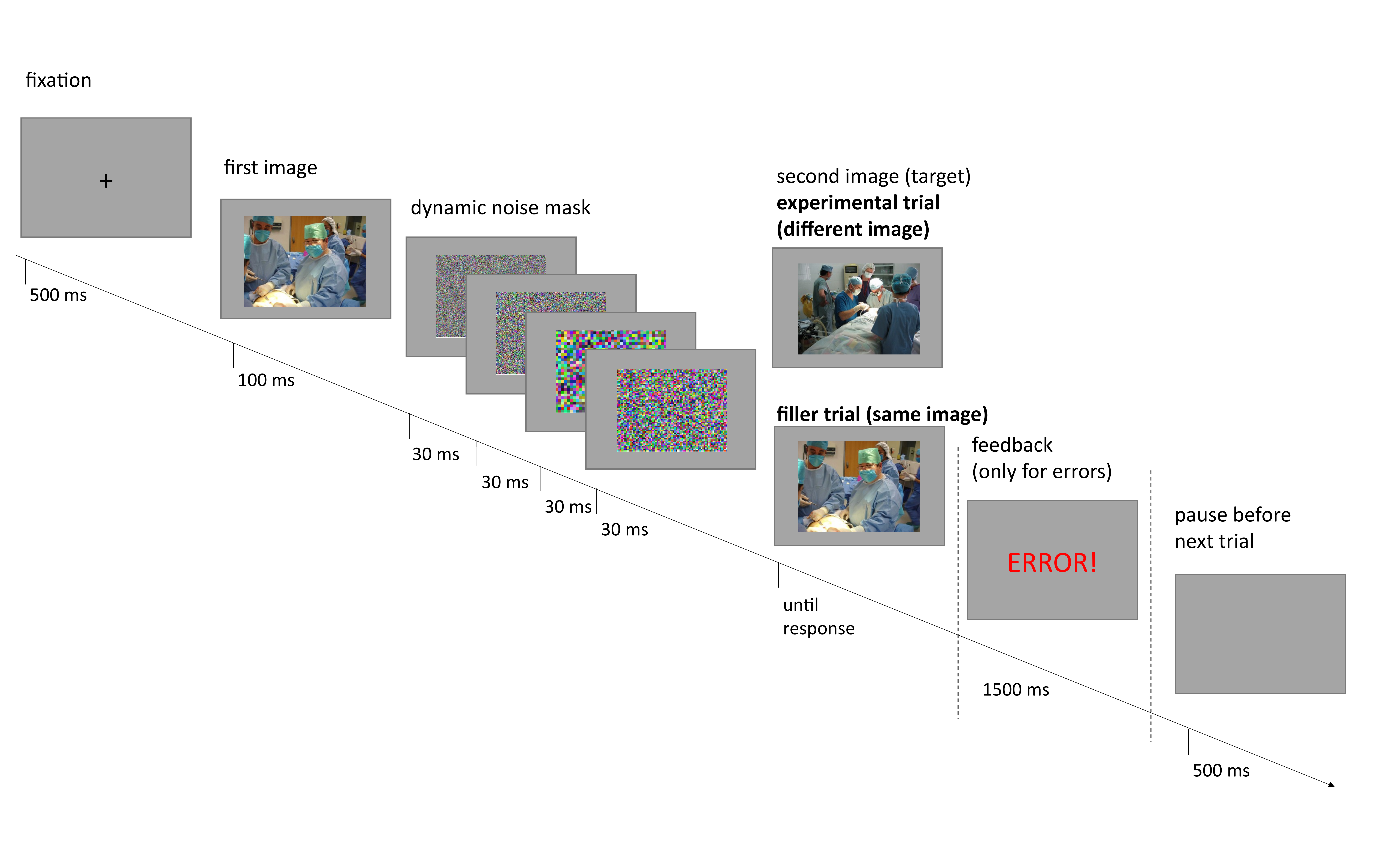}

\textbf{[PRIM]}

\includegraphics[width=\linewidth]{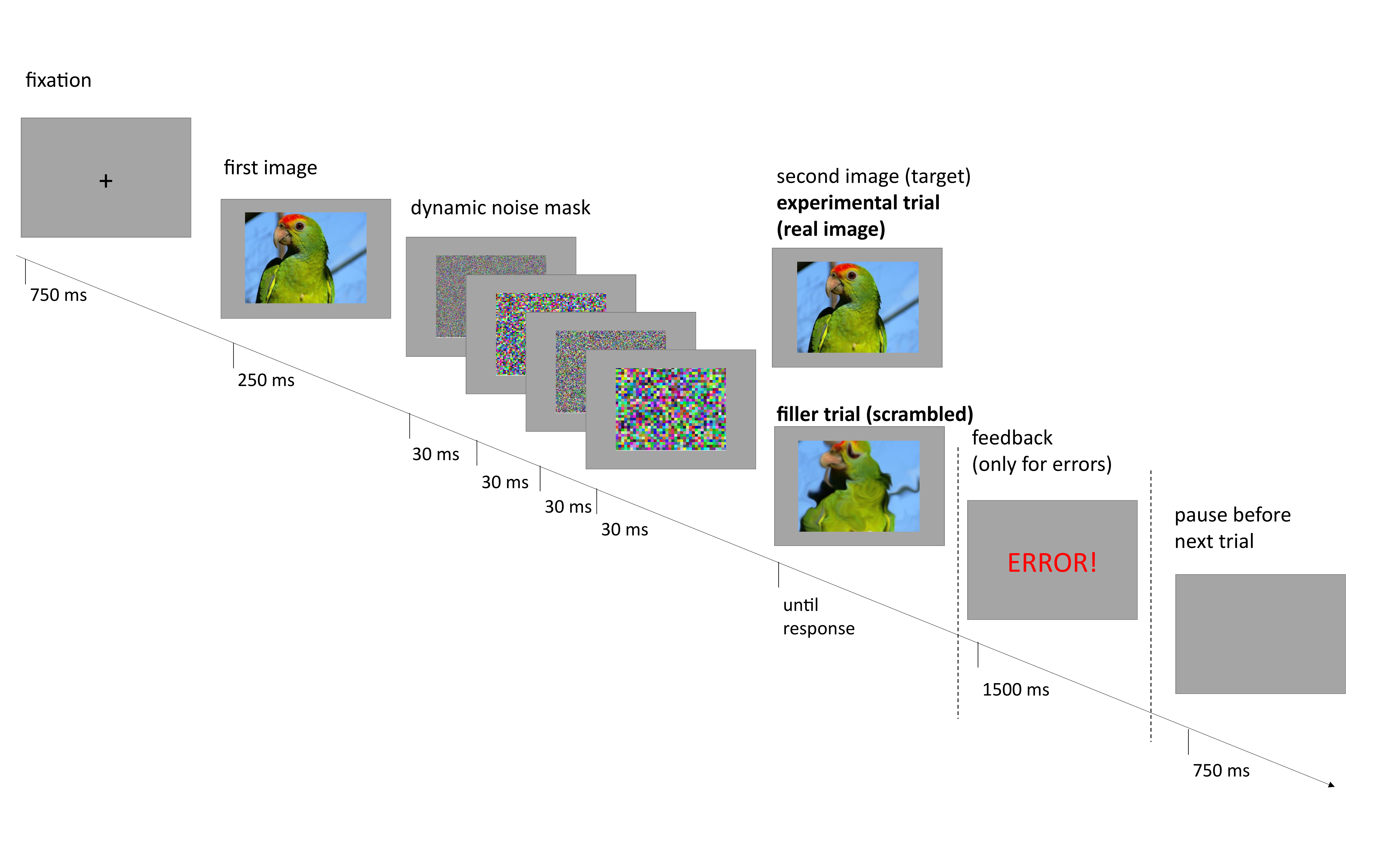}

\caption{Schematic representations of experimental trials in the processing time paradigms. \emph{Upper panel:} the discrimination task [DIS], in which participants have to decide whether the second image (the target) is identical to the first \emph{lower panel:} priming task [PRIM], where participants have to decide whether the second image (the target) is a real image or a scrambled one. The behavioral variable of interest is the time until a response is made for the target.}

\label{fig:trials}

\end{figure}

We considered the following metrics (see \cite{vispa} for detailed descriptions of the data collection procedures):

\begin{itemize}

\item \textbf{Ratings} of
\begin{itemize}
\item \textbf{image similarity [IMG]} for 3,000 pairs of naturalistic ImageNet images. Data were collected from 480 participants, with 30 observations per image pair.

\item \textbf{\emph{visual} word similarity [WORD]} for 3,000 word pairs (image labels of the aforementioned 3,000 image pairs), where participants were asked to judge how similar of the objects denoted by the words (i.e., the word referents) \emph{look like}. Thus, unlike other word-based ratings, \cite{Bruni2014,Finkelstein2001,Simlex}, these data focus on the visual domain. Data were collected from 480 participants, with 30 observations per word pair.

\item \textbf{typicality ratings [TYP]} for 7,500 word-image pairs (1,500 sets of image labels and five images tagged with that label), where participants were asked to indicate the most and least typical image for the category denoted by the presented label. Data were collected from 902 participants, with 30 observations per word-image pair.

\end{itemize}

All ratings were collected using the best-worst method \cite{Hollis2018}, so participants were always presented with a set of stimuli and asked to pick the most and least relevant for the given task. Responses were then scored on a continuous scale using the Value learning algorithm \cite{Hollis2018}. As a result, the datasets contain exactly one rating score between 0 (completely dissimilar) and 1 (identical) for each word pair in the WORD dataset  and each image pair in the IMG dataset, and one score between 0 (very atypical) and 1 (very typical) for each word-image pair in the TYP dataset. Examples for items with very high and very low ratings are presented in Figure~\ref{fig:rating_examples}

\begin{figure}[ht]

\includegraphics[width=\linewidth]{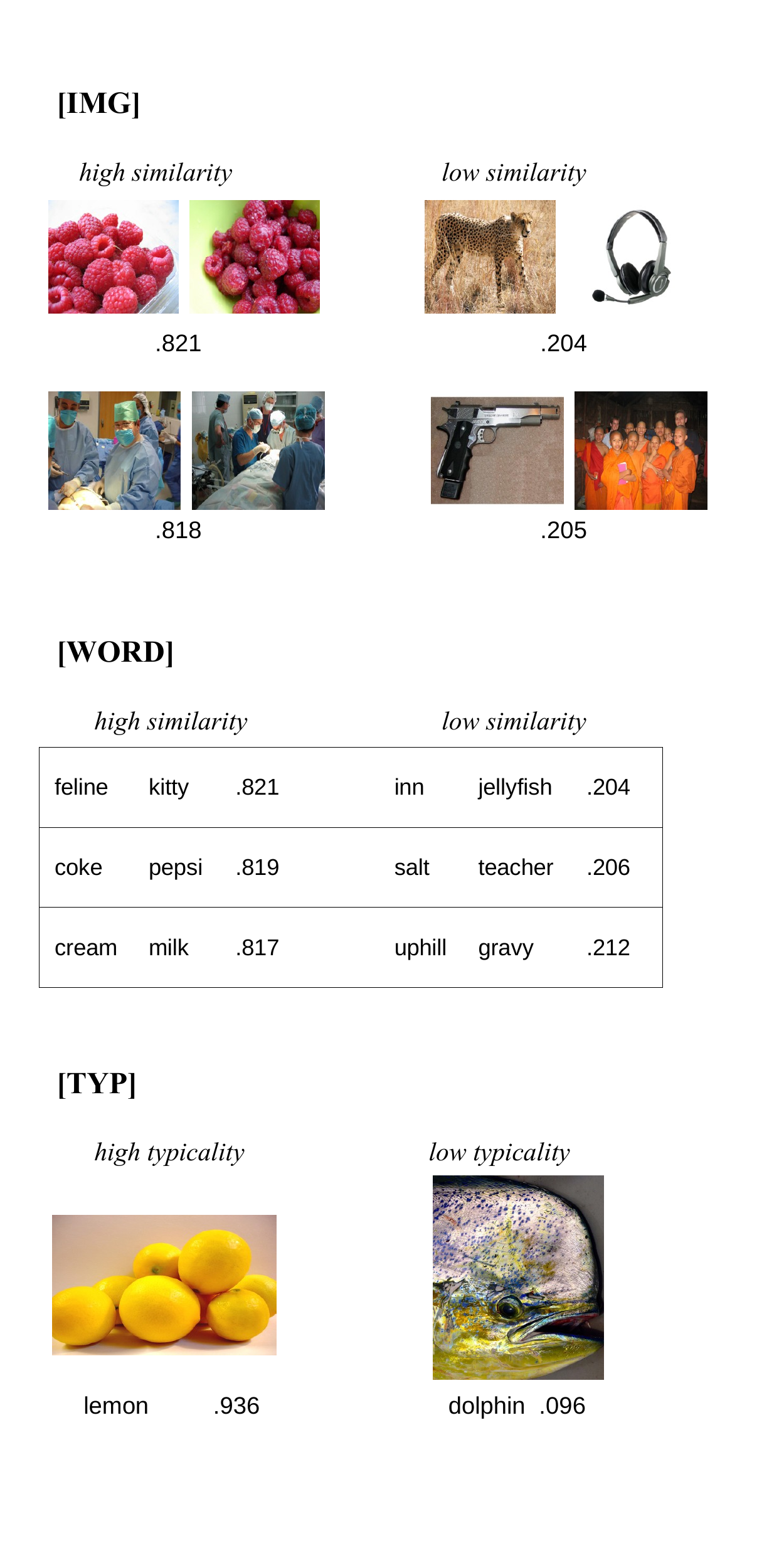}

\caption{Examples for items with very high and very low rating values in the individual rating tasks. \emph{Upper panel:} Image similarity ratings [IMG]; \emph{middle panel:} word similarity ratings for visual similarity between the words denoted by the objects [WORD], \emph{lower panel:} typicality ratings.}

\label{fig:rating_examples}

\end{figure}

\item \textbf{Processing time data}
\begin{itemize}
\item \textbf{discrimination task [DIS]}  for the same 3,000 image pairs of the IMG dataset. In a discrimination task, two stimuli (here: images) are presented in very rapid succession, and participants have to indicate whether they are identical or different by pressing one of two buttons (see Figure~\ref{fig:trials}, upper panel for a schematic representation of an experimental trial). Responses are typically \emph{slower} for more visually similar stimuli, which are harder to discriminate from the actual stimulus. Data were collected from 750 participants, with 30 observations per image pair.

\item \textbf{priming task [PRIM]} for the same 3,000 image pairs of the IMG and DIS datasets. In a priming task, two stimuli (here: images) are presented in quick succession, and participants have to perform a task on the second image only (here: judge whether a real or scrambled image has been presented by pressing one of two buttons); see Figure~\ref{fig:trials}, lower panel for a schematic representation of an experimental trial. Responses are typically \emph{faster} when the stimulus was preceded by a more visually similar stimulus, which primes (= facilitates processing of) the target. Data were collected from 750 participants, with 30 observations per image pair.

\end{itemize}

\end{itemize}

The target variable in these processing time studies is the mean response time for each image pair, after removing erroneous trials and outliers with far too slow or fast responses.

All datasets are publicly available in an OSF repository associated to the original study \cite{vispa} at  {\url{https://doi.org/10.17605/OSF.IO/QVW9C}}.

\section{Vision Models}

\subsection{Models employed}

For this study, we considered all pre-trained vision models available in the \emph{MatConvNet} \cite{MatConvNet} and \emph{Deep Learning Toolbox} (\url{https://github.com/matlab-deep-learning/MATLAB-Deep-Learning-Model-Hub}) packages for MATLAB. A full list of models is provided in Table~\ref{tab:models}.

\begin{table*}[]
    \centering
    \begin{tabular}{lrrrrr}
    \hline
       model  & \# layers & param. (mio.) & acc. & year & ref.  \\
 
        \hline
                AlexNet & 8 &  61.0 &  57.4 & 2012 & \cite{Krizhevsky2012} \\
                CaffeNet & 8  &   61.0  &  	57.4 & 2014 & \cite{Caffe} \\
                DarkNet-19 & 19 & 	20.8 & 74.0 & 2017 & \cite{Redmon2017} \\
                DarkNet-53 & 53 &  41.6 & 76.5 & 2017 & \cite{Redmon2017} \\
                DenseNet-201 & 201 & 20.0 & 75.9 & 2017 & \cite{DenseNet} \\
                EfficientNet B0 & 82 & 5.3 &  74.7    &  2019 & \cite{EfficientNet} \\
                GoogLeNet & 22 &  7.0 & 66.3  & 2015 & \cite{Szegedy2015} \\
                Inception-ResNet-v2 & 164 &  55.9 & 79.6  & 2017 & \cite{Szegedy2017}\\
                Inception-v3 & 48 & 23.9 & 77.1  & 2016 & \cite{Szegedy2016} \\
                MobileNetV2 & 53 & 	3.5 & 70.4    & 2018 & \cite{MobileNet} \\
                NASNet-Mobile & *  &  5.3 & 73.4 & 2018 & \cite{Zoph2018} \\
                ResNet-18 & 18 &    11.7 &   69.5    & 2016 & \cite{He2016} \\
                ResNet-50 & 50 & 25.6 &  74.5   & 2016  & \cite{He2016} \\
                ResNet-101 & 101 & 	44.6 &  76.0   & 2016 & \cite{He2016} \\
                ResNet-152 & 152 & 	60.3 &  77.0   & 2016 & \cite{He2016} \\
                ShuffleNet & 50 &  1.4 &  63.7  & 2018 & \cite{ShuffleNet} \\
                SqueezeNet & 18 & 1.2 & 55.2    & 2016 & \cite{SqueezeNet} \\            
                VGG-16 & 16  & 138.3 & 71.5   & 2014 & \cite{Simonyan2014} \\
                VGG-19 & 19  &  143.7 & 71.3 & 2014 & \cite{Simonyan2014} \\
                VGG-F & 8 & 60.8   &  58.9  & 2014 & \cite{Chatfield2014}  \\
                VGG-M & 8 & 102.9   &  62.7  & 2014 & \cite{Chatfield2014} \\
                VGG-M-128 & 8  & 82.7   &  59.2  & 2014 & \cite{Chatfield2014}  \\
                VGG-M-1024 & 8  &  87.2 & 62.2  &  2014 & \cite{Chatfield2014}  \\
                VGG-M-2048 & 8  &  92.5  &  62.9 & 2014 & \cite{Chatfield2014}  \\
                VGG-S & 8 & 102.9  & 63.3   &  2014 & \cite{Chatfield2014}  \\              
                Xception & 71 & 22.9 &  78.2 & 2017 & \cite{Chollet2017} \\

                \hline
    \end{tabular}
    \caption{Overview over the models investigated, including their number of layers, number of parameters, accuracy (measured as top-1 accuracy in the ImageNet classification task ILSVRC2012), and references to the papers introducing the models. \\
    *NASNet-Mobile does not consist of a linear sequence of modules}
    
    \label{tab:models}
    
\end{table*}


\subsection{General setup: Image and prototype representations}

In line with previous studies \cite{Battleday2020,BattledayReview,vispa,Petilli}, we extracted the activation values in each convolutional and fully-connected layer of a model for a given input image (i.e., image embeddings) as representations for that image. In addition, we constructed prototype vectors for image labels as the centroid of 100--200 image embeddings of images tagged with that label (using the very same method presented in \cite{vispa,Petilli}). For each image label, we obtain such a prototype representation for each layer of each considered model.

We used the cosine similarity metric to compute similarities between these image embeddings (at the same layer of the same model). In this manner, we can obtain a metric for the similarity between two individual images (for the IMG, DIS and PRIM datasets), the overall visual similarity between two categories denoted by their respective image labels (for the WORD dataset), and the similarity between an individual image and its category (for the TYP dataset) at each layer for each model.

\section{Results}

Since relations between the model-derived similarities and the behavioral outcome variables are mostly non-linear \cite{vispa}, performance was assessed using Spearman rank correlations. All predictors (i.e., similarities based on each layer of each model) were ranked in terms of performance on each behavioral dataset, and these ranks were used to calculate three general-level evaluation metrics: 

\begin{itemize}

    \item The \textbf{rating performance} as the mean rank across the three rating datasets (IMG, WORD, and TYP)

    \item The \textbf{processing time performance} as the mean rank across the two processing-time datasets (DIS and PRIM)

        \item The \textbf{overall performance} as the mean rank across all behavioral datasets  (compare \cite{Baronipredict,Gupta2021})

\end{itemize}
    
The results for the best-performing layers for each evaluation metric are displayed in Table~\ref{tab:resultstable}. We include the best-performing layer in the paper by G\"unther et al. \cite{vispa} (VGG-F, fully-connected layer 6) as a reference condition. Note that, for the PRIM dataset, participants tend to respond \emph{faster} (that is, lower response times) if the two images are more similar; therefore, the target metric here is a \emph{more negative} correlation.

\begin{table*}

\begin{center}

\footnotesize

\bgroup
\def\arraystretch{1.7}
\begin{tabular}{lll|rrrrr|rr|r}

\hline
row & model & layer & IMG & WORD & TYP & DIS & PRIM & rating & processing & overall \\
\hline

1 & GoogLeNet & 5a\_3x3\_reduce & \textbf{1 (0.774)} & \textbf{1 (0.666)} & 35 (0.361) & 65 (0.207) & 88 (-0.088) & 12.3 & 76.5  & \textbf{38.0}  \\

2 & DarkNet-19 & conv14 & 9 (0.740) & 81 (0.646) & 39 (0.359) & 43 (0.211) & 43 (-0.095) & 43.0 & 43.0  & 43.0  \\

3 & GoogLeNet & 5a\_3x3 &  41 (0.707) & 53 (0.649) & 95 (0.339) & 42 (0.211) & 11 (-0.105) & 63.0 & 26.5  & 48.4  \\
\hdashline
\\

4 & VGG-M-1024 & fc7 & 8 (0.741) & 11 (0.660) & 4 (0.389) & 137 (0.199) & 383 (-0.066) & \textbf{7.7} & 260.0  & 108.6  \\

5 & VGG-M-2048 & fc7 & 15 (0.734) & 21 (0.657) & 3 (0.392) & 185 (0.192) & 301 (-0.071) & 13.0 & 243.0  & 105.0  \\

\hdashline

6 & VGG-M & fc6 & 36 (0.714) & 288 (0.626) & 147 (0.324) & 12 (0.222) & 8 (-0.107) & 157.0 & \textbf{10.0}  & 98.2  \\ 

\hdashline 
\\

7 &DarkNet-53 & conv50 & 114 (0.648) & 17 (0.658) & \textbf{1 (0.400)} & 328 (0.176) & 56 (-0.092) & 44.0 & 192.0  & 103.2  \\

\hdashline 

8 & EfficientNet B2 & B12-D-conv2d-D* & 47 (0.702) & 42 (0.651) & 160 (0.322) & \textbf{1 (0.231)} & 129 (-0.083) & 83 & 65  & 75.8  \\ 

\hdashline 

9 & ResNet-50 & Res5a-Branch2b & 76 (0.679) & 113 (0.643) & 134 (0.327) & 80 (0.205) & \textbf{1 (-0.128)} & 107.7 & 40.5  & 80.8  \\

\hdashline 

\\

10 & VGG-F & fc6 & 24 (0.721) & 244 (0.63) & 176 (0.316) & 7 (0.224) & 18 (-0.102) & 148 & 12.5  & 93.8  \\

\hline

\end{tabular}
\egroup

\end{center}

\caption{The best-performing models across the different datasets, arranged by overall performance (the three top models in rows 1--3), rating performance (the two top models in rows 4--5), processing time performance (the top model in row 6), and performance in the individual datasets (the top models in row 1, as well as rows 7--10). The first number indicates the rank (ranging from a top value of 1 to a worst value of ), the second number in brackets the Spearman correlation. The best-performing model in G\"unther et al. \cite{vispa} (VGG-F, layer fc6)  is listed as a baseline (row 10). \\
*layer blocks-12-depthwise-conv2d-depthwise}

\label{tab:resultstable}

\end{table*}

As can be seen in Table~\ref{tab:resultstable}, the overall best-performing representations (i.e., the model estimates that are most associated with behavioral variables) are provided by the GoogLeNet model, more specifically one of the representations in the 5th layer of the model (5a\_3x3\_reduce). These representations are also best-performing when it comes to predicting the arguably most fundamental types of behavioral data, similarity judgments within a given modality (i.e., between two different images [IMG] and between two different categories [WORD]).

Focusing only on the explicit rating data ([IMG], [WORD], and [TYP]), the best-performing representations are provided by the 7th layer (a fully-connected layer) of the VGG-M-1024 variant, closely followed by the same layer of the VGG-M-2048 variant. Although these perform slightly worse for the [IMG] dataset than the best-performing GoogLeNet layer (and very marginally worse for the [WORD] dataset), they make up for this with a near top-level performance in the [TYP] dataset (with the 50th convolutional layer of the DarkNet-53 model as the top-performer). However, these models fall behind a mean rank of 240 for the processing time data.

When focusing only on the processing time data ([DIS] and [PRIM]), the 6th layer (again, a fully-connected layer) of the VGG-M model (standard variant) performs best, with near top-level performance in both individual datasets (those top-performers being a layer of the EfficientNet B2 model and of the ResNet-50 model, respectively). However, conversely to the 7th layers in the VGG-M-1024 and VGG-M-2048 variants, these representations in turn fall behind for the rating data, with a mean rank of 157.

\subsection{Comparison with model characteristics}

In an additional step, we assessed the relation between the characteristics of a model (more specifically, their number of parameters and their top-1 classification accuracy \cite{Russakovsky2015}; see Table~\ref{tab:models}) and its performance on the behavioral datasets tested here. To this end, we equated the overall model performance with the performance of its best-performing layer, as measured by the mean rank.

\begin{figure}[ht]

\begin{center}

\includegraphics[width=.4\textwidth]{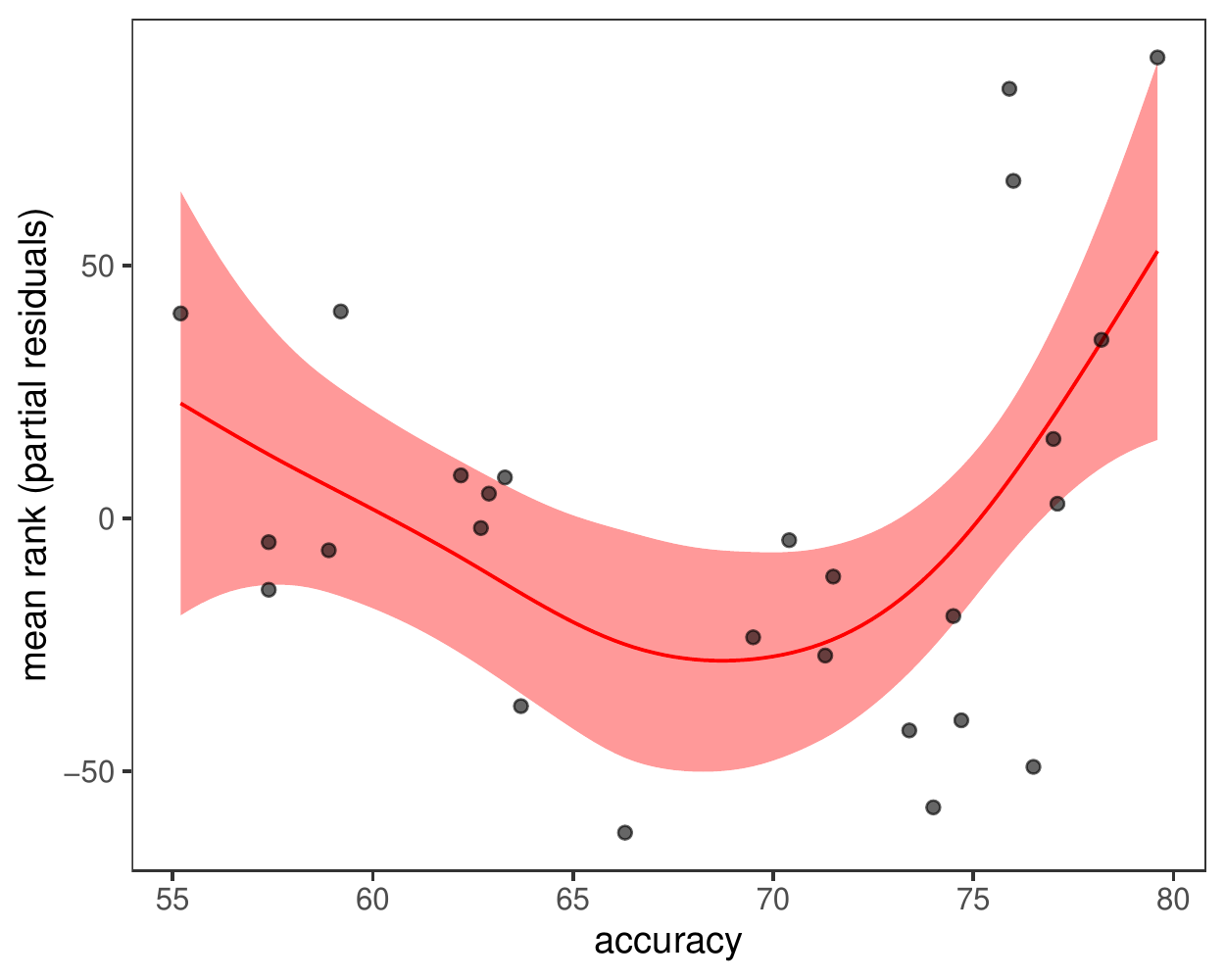}

\vspace{5mm}

\includegraphics[width=.4\textwidth]{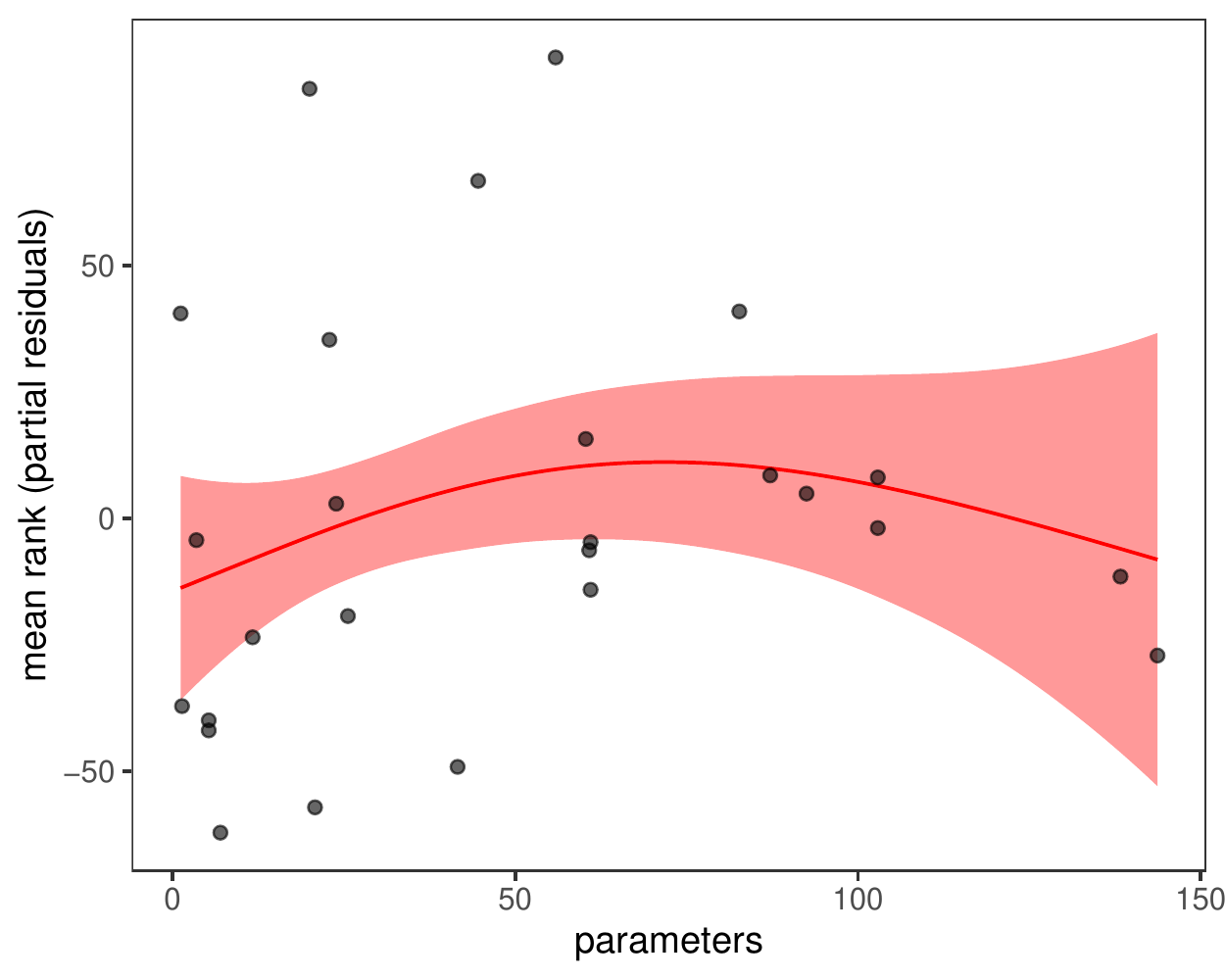}

\end{center}

\caption{The relation between a model's accuracy (upper panel) and number of parameters (lower panel) on the performance across all beahvioral datasets (measured as mean rank; the graphs show the partial residuals of a GAM analysis of this outcome variable). Each individual data point represents the best-performing layer of one of the models tested here. }

\label{fig:GAM}

\end{figure}

We estimated two separate non-linear statistical models (GAMs; \cite{mgcv,mgcViz}), modelling mean rank as a function of model accuracy and number of parameters, the results of which are depicted in Figure~\ref{fig:GAM}. Note that a \emph{lower} mean rank indicates better performance. As can be seen in these plots, medium levels of classification accuracy tend to be associated with better performance against behavioral data (with two local minima around 65\% and around 75\%). Regarding the number of parameters, either very low or very high numbers tend to be associated with better model performance; however, a closer inspection of the individual data points indicates that the best-performing models all have a rather low number of parameters.

\section{Discussion}

\subsection{Implications of the results}

In the present study, we investigated which representations obtained from different computer-vision models best predict a battery of five large-scale behavioral datasets, including both rating data and processing time data. We find that the overall best-performing models are in fact quite ``old'' models, given the pace of the research cycle within the field: A layer of the GoogLeNet model (published 2015, \cite{Szegedy2015}) displays the overall highest performance across all five datasets, and different layers (of different variants) of the VGG-M model (published 2014, \cite{Chatfield2014}) display the best overall performance for the rating data and processing time data, respectively. Note that the differences in performance between the individual representations are meaningful and not trivial: For example, the difference in performance for the [IMG] dataset between the overall best GoogLeNet layer (0.774) and the overall second-best DarkNet-19 layer (0.740) is already more than three percentage points.

Over the last years, a lot of effort has gone into developing systems with ever better performance than these ``older'' models. With respect to the task these models are designed for -- most prominently, image classification -- this effort has reached impressive successes: As can be seen in Table~\ref{tab:models}, the top-1 accuracy for the ILSVRC 2012 validation data \cite{Russakovsky2015} has increased dramatically, from around 60\% in 2014/2015 to around 80\%. In comparison, GoogLeNet (66.3 \%) and especially VGG-M (around 60 \% for all variants) definitely fall on the lower end of this scale. This however reveals an interesting rift opening with respect to model performance: Even though more recent models get better and better on their target tasks, this improvement in classification accuracy does not go along with improvements in predicting other types of data  (in fact the contrary, compare Figure~\ref{fig:GAM}, upper panel). This is not to say that more recent models show low performance on this type of data: Representations from recent models and highly accurate models like DarkNet-19 \cite{Redmon2017} are among the best-performing representations available. The critical point however still remains that the strong improvement in classification accuracy has not been accompanied by an \emph{improvement} in predicting other types of data.

On the other hand, we find no clear connection between model complexity and top performance in the behavioral dataset: The GoogLeNet model is relatively small in terms of parameters (7 mio.) and comes with an intermediate number of layers (22), while the VGG-M models are quite large  (around 90 to 100 mio. parameters) but have only a few layers (8). Therefore, one can neither conclude that a model needs to be very large and complex for top-level performance on behavioral data (consider especially the better performance of the  VGG-M model vis-\`a-vis the conceptually and architecturally similar VGG-16 and VGG-19 models), nor that it needs to be particularly small and efficient (compare also Figure~\ref{fig:GAM}, lower panel).

At this point, we can only speculate \emph{why} more recent and more classification-accurate models don't perform better in accounting for behavioral data. One possible explanation may be that the models are optimized to predict a very \emph{specific} image class, and only exact matches as a hit when calculating accuracy -- with the mis-classification of a \emph{spotted salamander} as a \emph{European fire salamander} treated as a miss in the same way a mis-classification as a  \emph{toaster} is. This may lead the models to weight relatively specific details to a similar or maybe even larger degree than the overall structure/``gestalt'' of the depicted object. Human judgments and responses, on the other hand, are more driven by these general-level similarities (e.g. \cite{Hebart2020}) rather than details (even if those are very informative for classification); this might lead to the observed discrepancy between classification accuracy and performance on behavioral datasets. However, we want to stress again that this is speculation on an open question, and more research is necessary to properly investigate and explain this discrepancy.

\subsection{Limitations and future directions}

At this point, we need to emphasize that all the issues discussed so far are based on the results of our evaluation, and therefore necessarily restricted to the models analysed here. However, there may well be models we did not consider here which contradict our findings (i.e., a high-accuracy model that simultaneously has a higher performance on behavioral data than the best-performing models identified here). In fact, in the context of successful transfer learning, we would consider this highly desirable, and hope that our study can give an impetus to systematically consider behavioral data in the search for an overall well-performing model.

While one may dismiss the behavioral data analysed here as not relevant for evaluating the performance of computer vision models, we argue that at the very least recognizing which images are more or less similar to one another should be considered one of the core prerequisites for a general-level vision model, analogous to semantic models predicting semantic similarity and relatedness data in the NLP domain \cite{Baronipredict,Bruni2014,Finkelstein2001,Simlex}.

In general, a desirable direction for future work in the field would be to develop general-level models that do not only excel in one particular task, but perform well across a range of different tasks. Ideally, in the spirit of successful transfer learning, this would not simply mean \emph{optimizing} a single model for a range of different tasks, but instead \emph{testing} such a model on a battery of tasks it was not optimized for \cite{imitationgame}. Following up on our suspicion that the lack of improvement in representation learning could be the result of hyper-engineering to distinguish very specific (and somewhat arbitrary) categories, we speculate that possible routes of advancement to achieve models representations that better capture a general similarity structure could be as follows: On the one hand, the training objective of the models could be altered to not only consider \emph{exact} hits among a set of candidate categories, but to also partially reward \emph{close} hits, for example based on their word embedding similarity or their WordNet distance to the correct target  (thus rewarding the classification of a \emph{poodle} as a \emph{dalmatian} or as a \emph{dog} more than as a \emph{Persian cat}, and that more than as a \emph{pillow}; see also \cite{DeDeyne2021}). On the other hand, the training sets of the models could be altered to more closely approximate human visual experience rather than over-representing certain categories \cite{Elgendy2020}, or to include more than one correct label per image \cite{Silberer2020}. 

We argue that such developments would be interesting from an engineering/transfer learning viewpoint (since a successful general-level model could be applied to new tasks that it was not originally optimized for), but also for the application of such systems as models of human visual representations in cognitive (neuro)science.

\section{Data availability}

Data and the analysis script for this study are available at \url{https://osf.io/sx5u3/?view_only=09c05b84a52246d5b8b061cbbee10350}.

{\small
\bibliographystyle{ieee_fullname}
\bibliography{lit}

\begin{thebibliography}{10}\itemsep=-1pt

\bibitem{Baronipredict}
Marco Baroni, Georgiana Dinu, and Germ{\'a}n Kruszewski.
\newblock {Don't count, predict! A systematic comparison of context-counting
  vs. context-predicting semantic vectors.}
\newblock In {\em {Proceedings of ACL 2014}}, pages 238--247, East Stroudsburg,
  PA, 2014. ACL.

\bibitem{Battleday2020}
Ruairidh~M Battleday, Joshua~C Peterson, and Thomas~L Griffiths.
\newblock Capturing human categorization of natural images by combining deep
  networks and cognitive models.
\newblock {\em Nature Communications}, 11:5418, 2020.

\bibitem{BattledayReview}
Ruairidh~M Battleday, Joshua~C Peterson, and Thomas~L Griffiths.
\newblock From convolutional neural networks to models of higher-level
  cognition (and back again).
\newblock {\em Annals of the New York Academy of Sciences}, 1505:55--78, 2021.

\bibitem{Bruni2014}
Elia Bruni, Nam-Khanh Tran, and Marco Baroni.
\newblock Multimodal distributional semantics.
\newblock {\em Journal of Artificial Intelligence Research}, 49:1--47, 2014.

\bibitem{Byerly2022}
Adam Byerly, Tatiana Kalganova, and Richard Ott.
\newblock The current state of the art in deep learning for image
  classification: A review.
\newblock In Kohei Arai, editor, {\em Intelligent Computing}, pages 88--105,
  Cham, 2022. Springer International Publishing.

\bibitem{Chatfield2014}
Ken Chatfield, Karen Simonyan, Andrea Vedaldi, and Andrew Zisserman.
\newblock Return of the devil in the details: Delving deep into convolutional
  nets.
\newblock {\em arXiv preprint; arXiv:1405.3531}, 2014.

\bibitem{Chollet2017}
Fran{\c{c}}ois Chollet.
\newblock Xception: Deep learning with depthwise separable convolutions.
\newblock In {\em Proceedings of the IEEE conference on computer vision and
  pattern recognition}, pages 1251--1258, 2017.

\bibitem{Cichy2019}
Radoslaw~M. Cichy and Daniel Kaiser.
\newblock Deep neural networks as scientific models.
\newblock {\em Trends in Cognitive Sciences}, 23:305--317, 2019.

\bibitem{DeDeyne2021}
Simon De~Deyne, Danielle~J Navarro, Guillem Collell, and Andrew Perfors.
\newblock Visual and affective multimodal models of word meaning in language
  and mind.
\newblock {\em Cognitive Science}, 45:e12922, 2021.

\bibitem{Elgendy2020}
Mohamed Elgendy.
\newblock {\em Deep Learning for Vision Systems}.
\newblock Manning, Shelter Island, NY, 2020.

\bibitem{mgcViz}
Matteo Fasiolo, Raphael Nedellec, Yannig Goude, and Simon~N. Wood.
\newblock {Scalable visualisation methods for modern Generalized Additive
  Models.}
\newblock {\em Arxiv preprint}, 2018.

\bibitem{Finkelstein2001}
Lev Finkelstein, Evgeniy Gabrilovich, Yossi Matias, Ehud Rivlin, Zach Solan,
  Gadi Wolfman, and Eytan Ruppin.
\newblock Placing search in context: The concept revisited.
\newblock {\em ACM Transactions on Information Systems}, 20:116--131, 2001.

\bibitem{TransferLearning}
Ian Goodfellow, Yoshua Bengio, and Aaron Courville.
\newblock {\em Representation Learning}, pages 524--554.
\newblock MIT press, 2016.

\bibitem{vispa}
Fritz G{\"u}nther, Marco Marelli, Sam Tureski, and Marco~Alessandro Petilli.
\newblock Vispa (vision spaces): A computer-vision-based representation system
  for individual images and concept prototypes, with large-scale evaluation.
\newblock {\em Psychological Review}, Advance online publication, 2022.

\bibitem{Gupta2021}
Abhijeet Gupta, Fritz G{\"u}nther, Ingo Plag, Laura Kallmeyer, and Stefan
  Conrad.
\newblock Combining text and vision in compound semantics: Towards a
  cognitively plausible multimodal model.
\newblock In {\em Proceedings of the 17th Conference on Natural Language
  Processing (KONVENS 2021)}, pages 218--222, 2021.

\bibitem{He2016}
Kaiming He, Xiangyu Zhang, Shaoqing Ren, and Jian Sun.
\newblock Deep residual learning for image recognition.
\newblock In {\em Proceedings of the IEEE Conference on Computer Vision and
  Pattern Recognition (CVPR 2016)}, pages 770--778, 2016.

\bibitem{Hebart2020}
Martin~N Hebart, Charles~Y Zheng, Francisco Pereira, and Chris~I Baker.
\newblock Revealing the multidimensional mental representations of natural
  objects underlying human similarity judgements.
\newblock {\em Nature Human Behaviour}, 4:1173--1185, 2020.

\bibitem{Simlex}
Felix Hill, Roi Reichart, and Anna Korhonen.
\newblock Simlex-999: Evaluating semantic models with (genuine) similarity
  estimation.
\newblock {\em Computational Linguistics}, 41:665--695, 2015.

\bibitem{Hollis2018}
Geoff Hollis.
\newblock Scoring best-worst data in unbalanced many-item designs, with
  applications to crowdsourcing semantic judgments.
\newblock {\em Behavior Research Methods}, 50:711--729, 2018.

\bibitem{DenseNet}
Gao Huang, Zhuang Liu, Laurens Van Der~Maaten, and Kilian~Q Weinberger.
\newblock Densely connected convolutional networks.
\newblock In {\em Proceedings of the IEEE conference on Computer Vision and
  Pattern Recognition (CVPR2017)}, pages 4700--4708, 2017.

\bibitem{SqueezeNet}
Forrest~N Iandola, Song Han, Matthew~W Moskewicz, Khalid Ashraf, William~J
  Dally, and Kurt Keutzer.
\newblock Squeezenet: Alexnet-level accuracy with 50x fewer parameters and< 0.5
  mb model size.
\newblock {\em arXiv preprint arXiv:1602.07360}, 2016.

\bibitem{Caffe}
Yangqing Jia, Evan Shelhamer, Jeff Donahue, Sergey Karayev, Jonathan Long, Ross
  Girshick, Sergio Guadarrama, and Trevor Darrell.
\newblock Caffe: Convolutional architecture for fast feature embedding.
\newblock {\em arXiv preprint arXiv:1408.5093}, 2014.

\bibitem{Jozwik2017}
Kamila~M Jozwik, Nikolaus Kriegeskorte, Katherine~R Storrs, and Marieke Mur.
\newblock Deep convolutional neural networks outperform feature-based but not
  categorical models in explaining object similarity judgments.
\newblock {\em Frontiers in psychology}, 8:1726, 2017.

\bibitem{Kriegeskorte2015}
Nikolaus Kriegeskorte.
\newblock Deep neural networks: a new framework for modeling biological vision
  and brain information processing.
\newblock {\em Annual Review of Vision Science}, 1:417--446, 2015.

\bibitem{Krizhevsky2012}
Alex Krizhevsky, Ilya Sutskever, and Geoffrey~E Hinton.
\newblock {ImageNet Classification with Deep Convolutional Neural Networks}.
\newblock In F. Pereira, C.~J.~C. Burges, L. Bottou, and K.~Q. Weinberger,
  editors, {\em {Advances in Neural Information Processing Systems 25}}, pages
  1097--1105, 2012.

\bibitem{Lindsay2021}
Grace~W Lindsay.
\newblock Convolutional neural networks as a model of the visual system: Past,
  present, and future.
\newblock {\em Journal of Cognitive Neuroscience}, 33:2017--2031, 2021.

\bibitem{Peterson2018}
Joshua~C Peterson, Joshua~T Abbott, and Thomas~L Griffiths.
\newblock Evaluating (and improving) the correspondence between deep neural
  networks and human representations.
\newblock {\em Cognitive Science}, 42:2648--2669, 2018.

\bibitem{Petilli}
M.~A. Petilli, F. G\"unther, A. Vergallito, M. Ciapparelli, and M. Marelli.
\newblock Data-driven computational models reveal perceptual simulation in word
  comprehension.
\newblock {\em Journal of Memory and Language}, 117:104194, 2021.

\bibitem{Redmon2017}
Joseph Redmon and Ali Farhadi.
\newblock Yolo9000: better, faster, stronger.
\newblock In {\em Proceedings of the IEEE conference on Computer Vision and
  Pattern Recognition (CVPR 2019)}, pages 7263--7271, 2017.

\bibitem{Roads2021}
Brett~D Roads and Bradley~C Love.
\newblock Enriching imagenet with human similarity judgments and psychological
  embeddings.
\newblock In {\em {Proceedings of the IEEE/CVF Conference on Computer Vision
  and Pattern Recognition}}, pages 3547--3557, 2021.

\bibitem{Russakovsky2015}
Olga Russakovsky, Jia Deng, Hao Su, Jonathan Krause, Sanjeev Satheesh, Sean Ma,
  Zhiheng Huang, Andrej Karpathy, Aditya Khosla, Michael Bernstein, et~al.
\newblock Imagenet large scale visual recognition challenge.
\newblock {\em International Journal of Computer Vision}, 115:211--252, 2015.

\bibitem{MobileNet}
Mark Sandler, Andrew Howard, Menglong Zhu, Andrey Zhmoginov, and Liang-Chieh
  Chen.
\newblock Mobilenetv2: Inverted residuals and linear bottlenecks.
\newblock In {\em Proceedings of the IEEE conference on Computer Vision and
  Pattern Recognition (CVPR 2018}, pages 4510--4520, 2018.

\bibitem{Silberer2020}
Carina Silberer, Sina Zarie{\ss}, and Gemma Boleda.
\newblock {Object Naming in Language and Vision: A Survey and a New Dataset}.
\newblock In {\em Proceedings of the 12th International Conference on Language
  Resources and Evaluation (LREC 2020)}, 2020.

\bibitem{Simonyan2014}
Karen Simonyan and Andrew Zisserman.
\newblock Very deep convolutional networks for large-scale image recognition.
\newblock {\em arXiv preprint, arXiv:1409.1556}, 2014.

\bibitem{imitationgame}
Aarohi Srivastava, Abhinav Rastogi, Abhishek Rao, Abu Awal~Md Shoeb, Abubakar
  Abid, Adam Fisch, Adam~R Brown, Adam Santoro, Aditya Gupta, Adri{\`a}
  Garriga-Alonso, et~al.
\newblock Beyond the imitation game: Quantifying and extrapolating the
  capabilities of language models.
\newblock {\em arXiv preprint arXiv:2206.04615}, 2022.

\bibitem{Szegedy2017}
Christian Szegedy, Sergey Ioffe, Vincent Vanhoucke, and Alexander Alemi.
\newblock Inception-v4, inception-resnet and the impact of residual connections
  on learning.
\newblock In {\em Proceedings of the AAAI conference on artificial
  intelligence}, volume~31, 2017.

\bibitem{Szegedy2015}
Christian Szegedy, Wei Liu, Yangqing Jia, Pierre Sermanet, Scott Reed, Dragomir
  Anguelov, Dumitru Erhan, Vincent Vanhoucke, and Andrew Rabinovich.
\newblock Going deeper with convolutions.
\newblock In {\em {Proceedings of the IEEE conference on computer vision and
  pattern recognition}}, 2015.

\bibitem{Szegedy2016}
Christian Szegedy, Vincent Vanhoucke, Sergey Ioffe, Jon Shlens, and Zbigniew
  Wojna.
\newblock Rethinking the inception architecture for computer vision.
\newblock In {\em Proceedings of the IEEE conference on computer vision and
  pattern recognition}, pages 2818--2826, 2016.

\bibitem{EfficientNet}
Mingxing Tan and Quoc Le.
\newblock Efficientnet: Rethinking model scaling for convolutional neural
  networks.
\newblock In {\em International Conference on Machine Learning}, pages
  6105--6114. PMLR, 2019.

\bibitem{MatConvNet}
Andrea Vedaldi and Karel Lenc.
\newblock {Matconvnet: Convolutional neural networks for Matlab}.
\newblock In {\em {Proceedings of the 23rd ACM international conference on
  Multimedia}}, pages 689--692. ACM, 2015.

\bibitem{mgcv}
Simon~N Wood.
\newblock {\em mgcv: Mixed {GAM} {C}omputation {V}ehicle with {GCV/AIC/REML
  S}moothness {E}stimation}, 2015.
\newblock R package version 1.8-7.

\bibitem{Zhang2018}
Richard Zhang, Phillip Isola, Alexei~A Efros, Eli Shechtman, and Oliver Wang.
\newblock The unreasonable effectiveness of deep features as a perceptual
  metric.
\newblock In {\em {Proceedings of the IEEE Conference on Computer Vision and
  Pattern Recognition}}, pages 586--595, 2018.

\bibitem{ShuffleNet}
Xiangyu Zhang, Xinyu Zhou, Mengxiao Lin, and Jian Sun.
\newblock Shufflenet: An extremely efficient convolutional neural network for
  mobile devices.
\newblock In {\em Proceedings of the IEEE Conference on Computer Vision and
  Pattern Recognition (CVPR 2018)}, pages 6848--6856, 2018.

\bibitem{Zoph2018}
Barret Zoph, Vijay Vasudevan, Jonathon Shlens, and Quoc~V Le.
\newblock Learning transferable architectures for scalable image recognition.
\newblock In {\em Proceedings of the IEEE conference on computer vision and
  pattern recognition}, pages 8697--8710, 2018.

\end{thebibliography}
}

\end{document}